\documentclass[letterpaper]{article} 
\usepackage{aaai2026}  
\usepackage{times}  
\usepackage{helvet}  
\usepackage{courier}  
\usepackage[hyphens]{url}  
\usepackage{graphicx} 
\usepackage{natbib}  
\usepackage{caption} 
\usepackage{booktabs}
\usepackage{algorithm}
\usepackage{algorithmic}
\usepackage{subfiles}
\usepackage{amsmath}
\usepackage{amsfonts}
\usepackage{nameref}

\usepackage{newfloat}
\usepackage{listings}
\DeclareCaptionStyle{ruled}{labelfont=normalfont,labelsep=colon,strut=off} 
\floatstyle{ruled}
\newfloat{listing}{tb}{lst}{}
\floatname{listing}{Listing}
\nocopyright
\title{SPANER: Shared Prompt Aligner\\ for Multimodal Semantic Representation}
\author{
    Thye Shan Ng, Soyeon Caren Han, Eun-Jung Holden
}
\affiliations{
    University of Melbourne

    thyeshan@student.unimelb.edu.au, caren.han@unimelb.edu.au, eunjung.holden@unimelb.edu.au
}

\usepackage{bibentry}

\begin{document}

\maketitle

\begin{abstract}
Recent advances in multimodal Parameter-Efficient Fine-Tuning (PEFT) have significantly improved performance on downstream tasks such as few-shot retrieval. However, most existing approaches focus on task-specific gains while neglecting the structure of the multimodal embedding space. As a result, modality-specific representations often remain isolated, limiting cross-modal generalisation.
In this work, we introduce Shared Prompt AligNER (SPANER), a modality-agnostic PEFT framework designed to embed inputs from diverse modalities into a unified semantic space. At its core, SPANER employs a shared prompt mechanism that acts as a conceptual anchor, enabling semantically related instances to converge spatially regardless of modality. This shared prompt design is inherently extensible, supporting the seamless integration of additional modalities, such as audio, without altering the core architecture.
Through comprehensive experiments across vision-language and audio-visual benchmarks, SPANER demonstrates competitive few-shot retrieval performance while preserving high semantic coherence in the learned embedding space. Our results highlight the importance of aligning embedding structures, rather than merely tuning adapter weights, for scalable multimodal learning.

\end{abstract}


\section{Introduction}
\begin{figure}[tbp]
    \centerline{\includegraphics[width=0.8\linewidth]{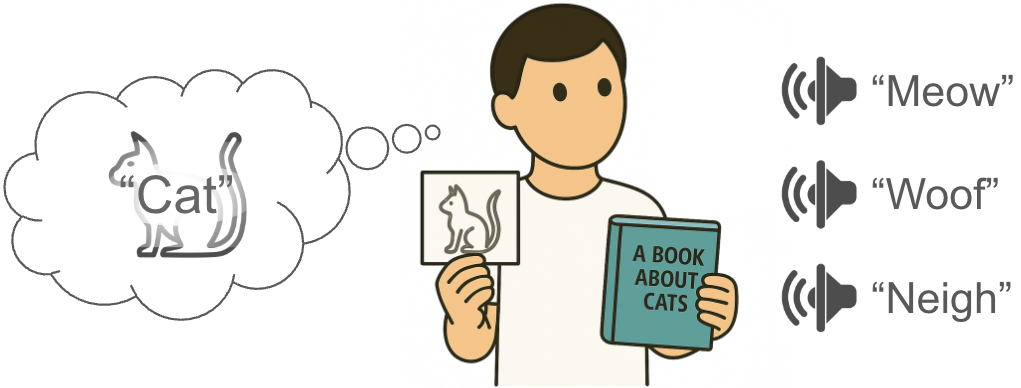}}
    \caption{A person forms the semantic concept of “cat” by combining information from different modalities--an image and a book. When presented with new audio cues, the person then decides which sound fits this concept. Our method enables such semantic grounding by aligning diverse modalities into a shared understanding space.}
    \label{fig:semantic_grounding}
    \end{figure}

Multimodal learning aims to unify diverse modalities, such as vision, language, and audio, into a shared semantic space that enables cross-modal understanding and inference \cite{ngiam2011multimodal}. Recent years have seen the emergence of powerful alignment models such as CLIP \cite{radford2021learning}, CLAP \cite{elizalde2023clap}, and their variants \cite{dou2022empirical, li2023scaling, wang2023image}, which are pretrained on large-scale multimodal data and enable tasks including image-text retrieval \cite{young2014image, deng2009imagenet}, visual question answering \cite{antol2015vqa, marino2019ok}, and audio-visual speech recognition \cite{son2017lip} through aligned feature representations.
While these alignment models demonstrate strong generalisation, their embeddings often remain modality-specific and are not explicitly optimised for semantic coherence across modalities. This limitation becomes more pronounced in fine-grained or domain-specific tasks, where modality gaps in the representation space may undermine performance \cite{liang2022mind, fahim2024s}. To address this, multimodal parameter-efficient fine-tuning (PEFT) strategies have emerged, which enable task-specific adaptation of pretrained alignment models by introducing lightweight, trainable components such as soft prompts \cite{zhou2022coop} or adapters \cite{gao2024clip}, while keeping the backbone frozen. These methods have led to significant improvements in few-shot retrieval and classification \cite{xing2024survey}, but their impact on the underlying structure of the embedding space remains underexplored.
Most existing multimodal PEFTs are evaluated solely on task-specific metrics. However, such evaluations often obscure whether the model actually improves semantic alignment or simply overfits to retrieval heuristics or prompt structures. As noted by prior work \cite{liang2022mind}, even alignment models trained with contrastive objectives tend to produce embeddings that reside in modality-specific cones or subspaces. This misalignment, known as the modality gap, restricts the model's ability to generalise across modalities and raises critical questions about the integrity of the shared semantic space. 
As illustrated in Figure~\ref{fig:semantic_grounding}, humans naturally integrate information from different modalities to form unified semantic concepts. Inspired by this intuition, we propose a novel framework that enables modality-agnostic semantic grounding through shared prompts and cross-modal alignment.

In this work, we propose SPANER (Shared Prompt AligNER), a modality-agnostic, plug-and-play PEFT framework designed to align all modalities into a coherent and extensible semantic space. At its core, SPANER introduces a shared prompt mechanism that serves as a conceptual anchor for modality-specific features, guiding their convergence in a unified embedding space. Unlike prior prompt-tuning approaches that apply soft prompts at the input level, SPANER’s shared prompt is applied post-encoder and fused via modality-specific cross-attention aligners. This architectural choice preserves the pretrained encoder’s strengths while enforcing semantic consistency across modalities.
To evaluate the semantic quality of learned embeddings, we complement standard task metrics with alignment-specific evaluations, such as average cosine similarity, modality-to-modality retrieval, and representation-level coherence measures. We conduct extensive experiments on vision-language tasks using CLIP backbones and extend our framework to the audio modality using EsResNeXt \cite{guzhov2021esresne}, following the design in AudioCLIP \cite{guzhov2022audioclip}. Across both settings, SPANER demonstrates strong few-shot retrieval performance and superior semantic alignment compared to prior PEFT methods.

\begin{itemize}
\item We propose \textbf{SPANER}, a modality-agnostic PEFT framework that aligns representations from diverse modalities into a unified semantic space via shared prompts and modality-specific aligners.

\item We introduce a \textbf{shared prompt mechanism} that acts as a conceptual semantic anchor, enabling extensible alignment across modalities without retraining on language input for new modalities.

\item We conduct comprehensive experiments across \textbf{vision-language and audio-visual tasks}, demonstrating that SPANER achieves competitive few-shot performance while significantly improving semantic alignment and reducing the modality gap.

\item We show that SPANER is \textbf{extensible to new modalities} (e.g., audio) in a plug-and-play manner, without altering the core architecture, highlighting its potential for scalable multimodal representation learning.
\end{itemize}

\section{Related Works}
The success of large-scale unimodal pretrained models \cite{devlin2019bert, dosovitskiy2020image, zhou2024comprehensive} has sparked a transition toward multimodal representation learning. Alignment models such as CLIP \cite{radford2021learning}, ALIGN \cite{jia2021scaling}, Florence \cite{yuan2021florence}, AudioCLIP \cite{guzhov2022audioclip}, and CLAP \cite{laionclap2023} aim to project inputs from multiple modalities (e.g., vision, language, and audio) into a shared representation space using contrastive learning objectives \cite{oord2018representation}. These models support a wide range of cross-modal tasks, including retrieval, captioning, and question answering.
While powerful, these models are typically optimised for general-purpose alignment and often underperform in fine-grained or domain-specific tasks. To address this, multimodal Parameter-Efficient Fine-Tuning (PEFT) methods have emerged, enabling efficient adaptation to downstream tasks by training a small number of additional parameters. Most existing multimodal PEFTs have been applied to vision-language (VL) settings using CLIP backbones \cite{zhou2022coop, zhou2022cocoop, shu2022test, gao2024clip, yang2024mma, khattak2023maple}, with more recent extensions to audio-language (AL) alignment in CLAP-based systems \cite{li2024audio, liang2023adapting}.
Broadly, multimodal PEFTs fall into two main categories: Prompt-based methods and Adapter-based methods.

\paragraph{Prompt-based PEFTs.}
Prompt-based approaches introduce learnable tokens at the input of frozen encoders, steering the model without modifying backbone weights. For instance, instead of using a fixed, manually crafted prompt such as $[This, is, an, image, of, a, [$CLASS$]]$, a set of learnable continuous prompt vectors is employed, such as $[V_1, ..., V_n,[$CLASS$]]$, where the weights of $V_i$ are updated during training. 
CoOp \cite{zhou2022coop} replaces handcrafted prompts with trainable vectors in the text encoder, while CoCoOp \cite{zhou2022cocoop} incorporates image context. Subsequent methods explore dynamic prompt updates during inference (TPT \cite{shu2022test}), bidirectional prompting \cite{wang2023tuning}, and prompt coupling across modalities (MaPLe \cite{khattak2023maple}). However, these prompts act only at the input level, and their influence fades as representations propagate through the frozen encoder. Consequently, the output embeddings still reflect modality-specific biases \cite{liang2022mind}, limiting the effectiveness of semantic alignment.


\paragraph{Adapter-based PEFTs.}
Adapter-based methods inject lightweight modules near the output of the encoders to refine task-specific features. For example, CLIP-Adapter \cite{gao2024clip} appends MLP layers after the image encoder, while Tip-Adapter \cite{zhang2021tip} performs few-shot adaptation via nearest-neighbor classification over cached features. MMA \cite{yang2024mma} inserts adapters within the top layers of CLIP’s encoders, assuming similar architectures across modalities. While effective, such methods tightly couple adaptation logic to specific backbones (e.g., Transformers for both vision and language), which restricts flexibility and generalisation, particularly when extending to heterogeneous architectures like CNN-based audio encoders.
Overall, existing multimodal PEFTs are often evaluated on downstream task performance, particularly few-shot classification, but rarely interrogate how well they align modality embeddings at the semantic level. Moreoever, prior methods are typically designed for dual-modality settings, making it difficult to scale to new modalities without architecture-specific engineering.

\paragraph{Our Contributions.}
In contrast to prior work, we focus on the structure of the shared embedding space rather than only downstream accuracy. We propose SPANER, a modular and extensible PEFT framework that unifies diverse modalities through a shared prompt and modality-specific cross-attention aligners. Unlike input-level prompts or adapter-specific modules, SPANER aligns representations post-encoder while preserving modality-agnostic compatibility. Our approach enables scalable semantic alignment across vision, language, and audio, without modifying pretrained backbones or requiring task-specific re-engineering.

\vspace{-1em}
\section{SPANER: Shared Prompt Aligner for Multimodal Semantic Representation}

Embedding inputs from diverse modalities, such as text, images, and audio, into a shared semantic space is a central goal in multimodal learning. This space enables semantically similar instances to be positioned nearby, facilitating cross-modal retrieval and understanding. However, structural differences between modalities (e.g., sequential text vs. spatial images) make this alignment inherently challenging \cite{baltruvsaitis2018multimodal}. Existing multimodal PEFTs often adopt direct interaction strategies that assume structural compatibility \cite{xing2024survey, gao2024clip, yang2024mma, khattak2023maple}, limiting scalability when extending to new modalities.

\begin{figure}[t]
    \centerline{\includegraphics[width=0.8\linewidth]{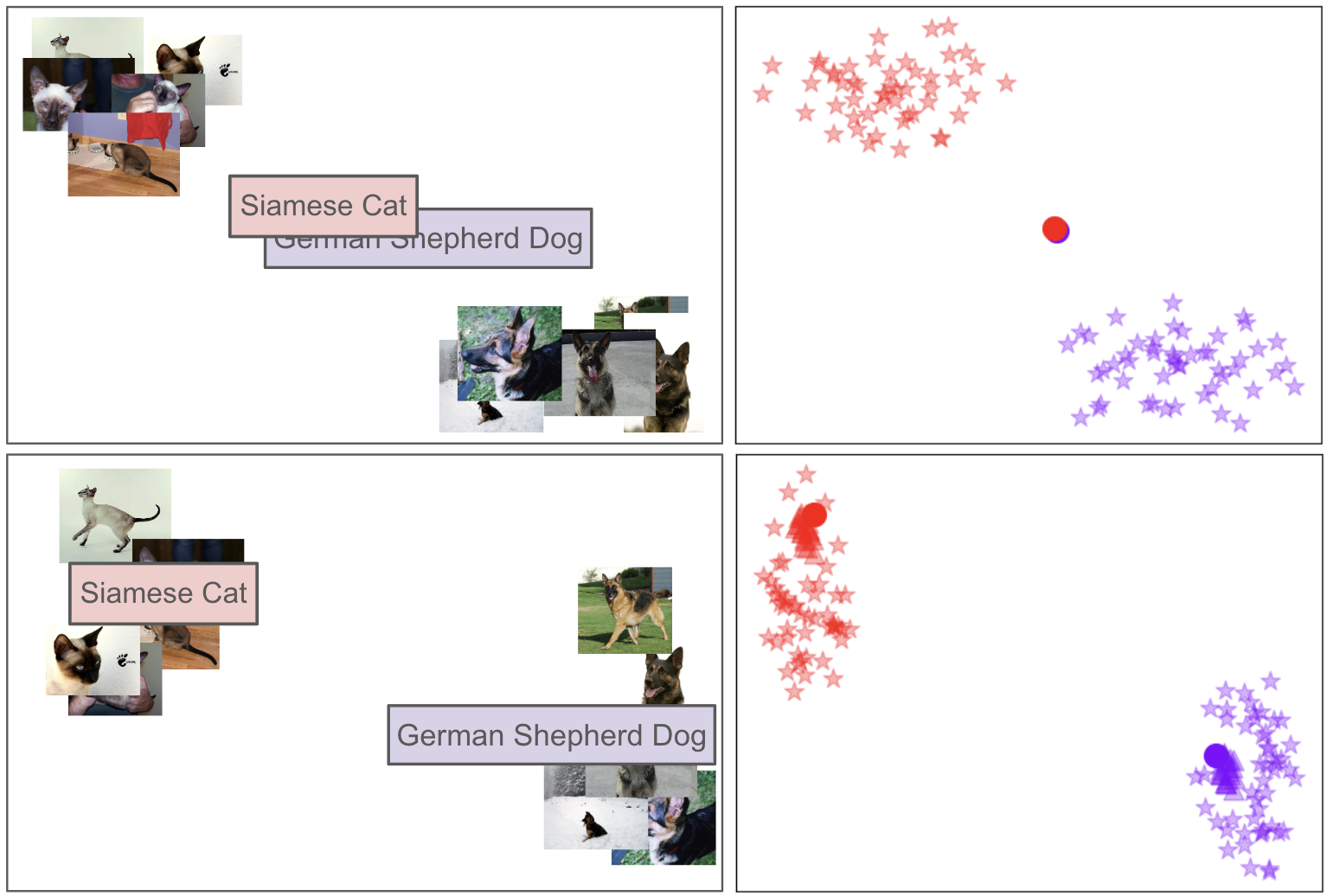}}
    \caption{t-SNE comparison of alignment quality between \cite{xing2024survey} (top) and our proposed method (bottom), with the plots showing the corresponding t-SNE of the embeddings, where $\star$ represents visual features, $\circ$ and $\triangle$ represent semantic and text of language features, respectively.}
    \label{fig:align}
    \end{figure}
\begin{figure*}[t]
    \centerline{\includegraphics[width=0.85\linewidth]{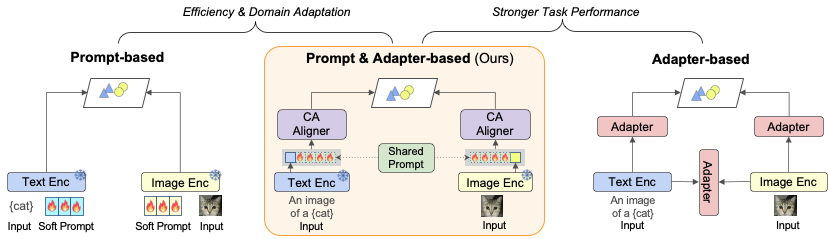}}
    \caption{Comparison of previous PEFT strategies with our proposed SPANER. \textbf{Prompt-based} methods append learnable vectors to the inputs of frozen encoders, guiding representations through input-level conditioning \cite{zhou2022coop,zhou2022cocoop,khattak2023maple}. \textbf{Adapter-based} approaches either insert lightweight modules at the output level of encoders \cite{gao2024clip,liang2023adapting,zhang2021tip}, or at the encoder level \cite{yang2024mma}, often involving modality-specific interactions. \textbf{Our method} leverages both methods by attaching a shared prompt and lightweight CA Aligner at the output level of encoders. The shared prompt serves as a modality-agnostic scaffold that guides both modalities toward a common semantic space, promoting alignment without requiring deep architectural coupling.
}
    \label{fig:design}
    \end{figure*}

To address this, we propose SPANER (Shared Prompt AligNER), a unified framework that aligns modality-specific representations through a shared semantic prompt (Figure~\ref{fig:design}). SPANER maps semantically equivalent instances into a common latent space using conceptual grounding signals—operationalised here as class labels. This design promotes modularity and extensibility, allowing new modalities to be integrated without retraining or architectural changes.
The following sections detail our SPANER for semantic alignment through shared prompts. 
We then illustrate its instantiations across different modality combinations, including vision, language, and audio. These applications underscore the flexibility and generalisability of our approach for aligning heterogeneous data within a unified semantic space.
In Figure~\ref{fig:align}(bottom), a semantically aligned space provides a natural scaffold to integrate new modalities, while unaligned modality-specific spaces in Figure~\ref{fig:align}(top) lack a meaningful reference, hindering scalability and generalisation.

\label{sec:methodology}




\subsubsection{Representation}
To establish this unified semantic space, we begin with inputs from two distinct modalities, each processed by its encoder to produce modality-specific embeddings, denoted \(x^{1}\) and \(x^{2}\). These initial embeddings reside in their respective representation spaces. To enable cross-modal alignment, we introduce two key components: a shared semantic prompt and a cross-attention (CA) aligner.

\subsubsection{Shared Prompt}
Departing from traditional soft prompts appended at the input layer \cite{zhou2022coop}, our shared prompt is introduced at the output level. It acts as a set of learnable conceptual tokens, functioning as semantic anchors to encourage alignment between modalities. Importantly, this prompt is shared across all modalities, ensuring consistent semantic grounding.
Each encoder output is augmented with the shared prompt tokens, resulting in combined representations of the form \([x^{1};S]\) and \([x^{2};S]\) where \(S=[s_1,\ldots,s_L]\) denotes the learnable prompt tokens. These tokens provide a common semantic context that guides subsequent alignment.

\subsubsection{Cross-Attention Aligner}  
The Cross-Attention Aligner (CA Aligner) processes each concatenated representation \([x^{i}; S]\), enabling information fusion between the modality-specific features \(x^{i}\) and the shared semantic prompt \(S\).  
Instead of allowing direct interactions between different modalities, each \(x^{i}\) attends only to \(S\), which acts as a semantic mediator. This design grounds each modality in a common semantic space while avoiding cross-modal entanglement.  
Fusion is performed using a standard attention mechanism with a residual connection~\cite{vaswani2017attention}, followed by a linear projection.  
Each modality is equipped with its own CA aligner, which is independently parameterised but structurally identical.
Formally, for each modality \(i\), the aligned representation \(f^{i}\) is computed as:
\vspace{-0.3em}
\begin{equation}\label{eq:attn_outs}
    [x^{i\prime}; S'] = \text{Attn}^{i}([x^{i}; S])
\end{equation}
\begin{equation}\label{eq:feature_outs}
    f^{i} = \text{Proj}^{i}(x^{i\prime} + x^{i}) 
\end{equation}
\vspace{-0.5em}
To optimise cross-modal alignment, we adopt a contrastive learning objective applied to the representations generated by the CA aligners. This objective is inspired by the CLIP-style formulation \cite{radford2021learning}, with full algorithmic details provided in Algorithm~\ref{alg:contrastive_loss} from the Appendix. Specifically, given a batch of size \(B\), and the aligned feature representations \(z^1\) and \(z^2\) from two modalities, we compute a pairwise cosine similarity matrix. The contrastive loss is then formulated as a cross-entropy loss over this similarity matrix, where each diagonal entry, corresponding to the matched pairs, is treated as the positive (ground truth) instance, and all off-diagonal entries as negatives. We apply this loss at two levels of the architecture:

(1) \textbf{CA alignment loss}: This loss is computed from the intermediate outputs of the CA aligner, retrieved from Equation~\ref{eq:attn_outs}. A max pooling operation is applied across the concatenated outputs to derive the aligned representations \(z^1\) and \(z^2\), and compute the contrastive loss: 
\begin{equation}
    \mathcal{L}_\text{CA} = \mathcal{L}(z^1,z^2)=\mathcal{L}(\text{Pool}([x^{1\prime};S^\prime]),\text{Pool}([x^{2\prime};S^\prime]))
\end{equation}

(2) \textbf{Final Alignment Loss}: This loss is based on the final projected embeddings \(f^1\) and \(f^2\), obtained from Equation~\ref{eq:feature_outs}, enforcing alignment in the final embedding space: \(\mathcal{L}_\text{align} = \mathcal{L}(z^1,z^2)=\mathcal{L}(f^1,f^2)\).

The overall training objective combines both losses, encouraging alignment both during intermediate processing and in the final representation space: \(\mathcal{L} = \mathcal{L}_\text{align} + \lambda\mathcal{L}_\text{CA}\), where \(\lambda\) is a hyperparameter controlling the relative contribution of the CA alignment loss.
The full training process is detailed in Algorithm~\ref{alg:shared_prompt_training} in Appendix.

\section{Toward Vision-Language Alignment}
\label{sec:vl}

In this section, we present the adaptation of the SPANER framework to the vision-language (VL) alignment task. We utilise the CLIP model \cite{radford2021learning} as a backbone, a standard choice in the multimodal PEFT literature, and augment it with our shared prompt and cross-attention alignment mechanisms to improve semantic consistency across image and text modalities.
Through a series of experiments on ImageNet \cite{deng2009imagenet}, we demonstrate the effectiveness of our approach, not only in retrieval performance but also in closing the gap between modality-specific representations and shared conceptual understanding. The following subsections detail CLIP integration, experimental setup, evaluation metrics, and key results.

\subsubsection{CLIP as a Backbone for SPANER}
SPANER is a modality-agnostic framework that operates on fixed-length feature representations from any encoder, denoted $x^{\text{img}}$, $x^{\text{text}}$, or more generally $x^i$ for modality $i$. For our vision-language experiments, we adopt CLIP \cite{radford2021learning}, specifically the ViT-B/16 variant as the backbone encoder, in line with prior work \cite{yang2024mma, gao2024clip, xing2024survey}.
CLIP consists of two transformer-based encoders for image and text, trained via contrastive learning on large-scale image-text pairs. For image inputs, visual tokens are extracted by projecting non-overlapping patches and prepending a [CLS] token. The output embedding at [CLS] serves as the image representation $x^{\text{img}}$. Text inputs are tokenised using Byte Pair Encoding, embedded, and appended with an [EOT] token, whose embedding is used as the text representation $x^{\text{text}}$.

SPANER integrates with CLIP by appending a shared prompt to each modality’s embedding and passing the result through modality-specific cross-attention aligners (see Section~\nameref{sec:methodology}). This enables semantically consistent alignment without modifying the pretrained CLIP encoders. The shared prompt and aligners structure the embedding space, allowing SPANER to enhance semantic coherence while preserving CLIP’s pretrained strengths.

\subsubsection{Experimental Setup}

We evaluate SPANER on ImageNet-1K under a 16-shot setting, following the CoOp protocol \cite{zhou2022coop} to ensure comparability with existing multimodal PEFT-based retrieval methods. Unlike approaches focused solely on classification, we prioritise semantic alignment and exclude benchmarks that do not reflect representational coherence.
During training, positive image-text pairs are constructed using randomly selected templates (e.g., ``an image of [CLASSNAME]'') from a predefined list \cite{lin2023crossmodal}. To avoid overfitting to prompt structure, we use only raw class labels (e.g., “golden retriever”) at test time for retrieval queries.
Training uses AdamW with a cosine annealing schedule, an initial learning rate of 0.001, weight decay of 0.01, and batch size of 256 for 30 epochs. The shared prompt is randomly initialised and optimised jointly with the contrastive objective, using a fixed temperature for consistency across runs.
We evaluate performance using top-1 retrieval accuracy for both text and semantic queries, alongside the average cosine similarity between matched pairs. Robustness under distributional shifts is assessed on ImageNet-V2, Sketch, A, and R variants (see Appendix for details), offering a comprehensive view of SPANER’s generalisability across visual domains. For qualitative analysis, a confusion matrix of semantic retrieval results is also provided in the Appendix to illustrate common misalignment patterns.

\subsubsection{Retrieval Accuracy vs. Semantic Coherence}

\begin{table}[]
    \centering
    \begin{tabular}{l|ccc}
    \hline
          \textbf{Method} & \textbf{Text} $(\uparrow)$ & \textbf{Semantic} $(\uparrow)$ & \textbf{Gap} $(\downarrow)$ \\
         \hline
         ZS CLIP & 66.7 & 64.1 & \underline{2.6} \\
         CoOp & \textbf{71.5} & 64.1 & 7.4 \\
         CLIP-Adapter & \underline{71.1} & 64.9 & 6.2 \\
         MMA & 71 & \underline{67.3} & 3.7 \\
         \textbf{SPANER (Ours)} & 69.7 & \textbf{67.6} & \textbf{2.1} \\
    \hline
    \end{tabular}
    \caption{Comparison of text vs. semantic retrieval accuracy on ImageNet (accuracy \%). Our method attains the smallest gap between text and semantic, suggesting a more semantically consistent embedding space without heavily sacrificing retrieval performance.
}
    \label{tab:itr_vs_semantic}
\end{table}
While retrieval accuracy is a common proxy for multimodal alignment, it does not always reflect semantic coherence. Many PEFT methods achieve high retrieval scores yet produce embeddings that remain modality-specific or overly reliant on prompt structures.
As shown in Table~\ref{tab:itr_vs_semantic}, CoOp attains the highest retrieval accuracy (71.5\%) but suffers from a large semantic alignment gap (7.4\%). Similar trends are seen in CLIP-Adapter (6.2\% gap) and MMA (3.7\%). These discrepancies indicate that high retrieval performance can result from surface-level matching rather than true cross-modal understanding.
In contrast, SPANER achieves slightly lower retrieval accuracy (69.7\%) but the highest semantic retrieval score (67.6\%), with only a 2.1\% gap, indicating stronger semantic alignment. We attribute this to SPANER’s shared prompt and cross-attention aligners, which enforce alignment at the representation level, reducing modality-specific drift and prompt overfitting.
These findings underscore the importance of evaluating alignment beyond retrieval metrics. SPANER better preserves semantic structure across modalities, making it more reliable for tasks requiring concept-level generalisation.



\begin{table}[]
    \centering
    \begin{tabular}{l|cc}
    \hline
         \textbf{Method} & \textbf{Text} $(\uparrow)$ & \textbf{Semantic} $(\uparrow)$ \\
     \hline
         ZS CLIP &  0.329 & 0.305 \\
         CoOp & \underline{0.339} & 0.305 \\
         CLIP-Adapter & 0.309 & 0.301 \\
         MMA & 0.321 & \underline{0.317} \\
         \textbf{SPANER} (Ours) & \textbf{0.748} & \textbf{0.691} \\
         
    \hline
         
    \end{tabular}
    \caption{Average cosine similarity scores of positive classes for text retrieval and semantic retrieval}
    \label{tab:cosineimage}
\end{table}

\begin{figure}[t]
    \centerline{\includegraphics[width=0.8\linewidth]{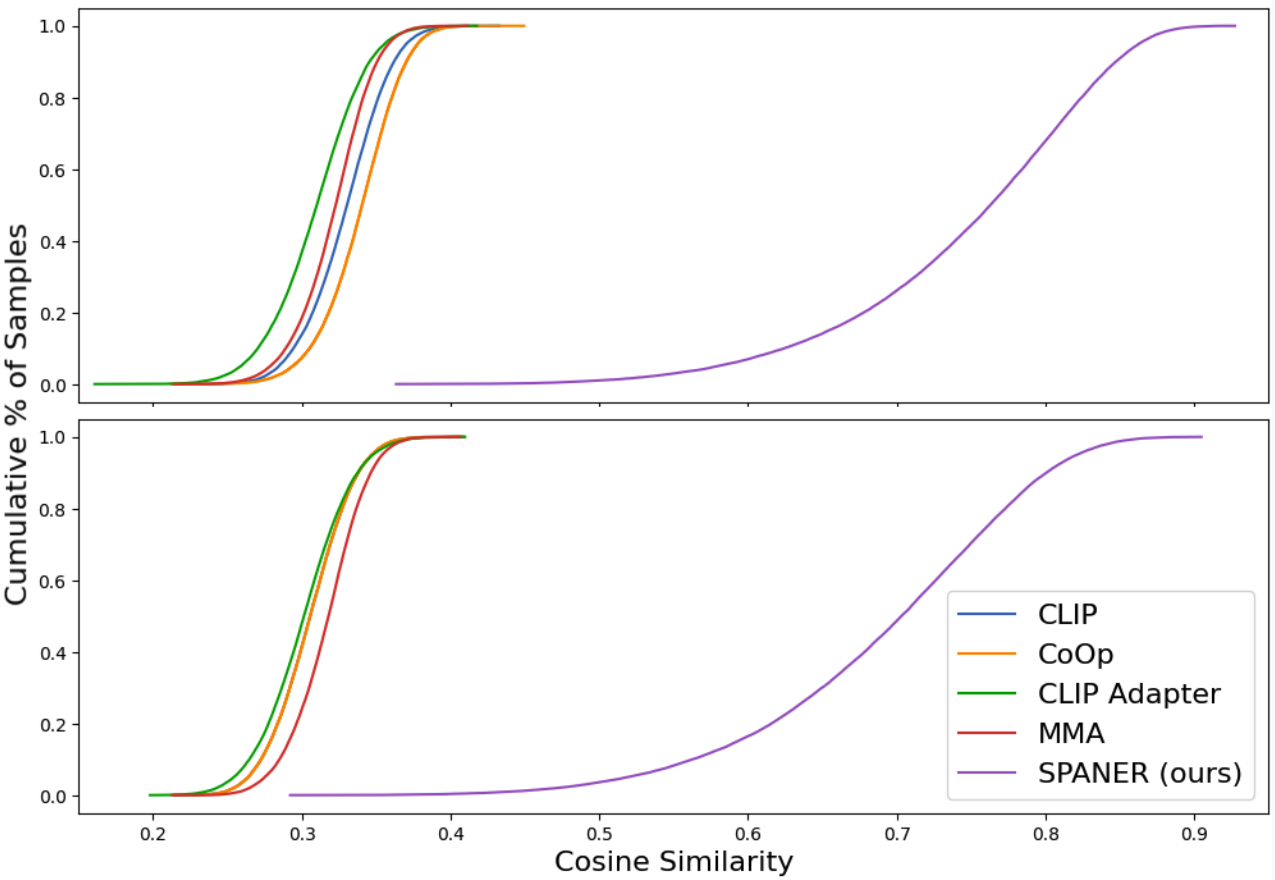}}
    \caption{Cumulative Distribution Function (CDF) of pairwise cosine scores of positive samples in ImageNet. Top: CDF for text retrieval, Bottom: CDF for semantic retrieval
    }
    \vspace{-0.3cm}
    \label{fig:cosineimage}
    \end{figure}

\subsubsection{Towards a Shared Semantic Space}
To assess the quality of semantic alignment, we compute the average cosine similarity between matched image-text pairs. Unlike retrieval accuracy, this metric reveals how tightly semantically equivalent items are clustered in the embedding space.
As shown in Table~\ref{tab:cosineimage}, SPANER significantly outperforms prior methods, achieving similarity scores of 0.748 (text) and 0.691 (semantic), compared to 0.30–0.33 for baselines like CoOp and MMA. This indicates that SPANER maps positive pairs into a more compact and coherent region of the shared space.

To visualise this clustering effect, we plot the cumulative distribution function (CDF) of cosine similarity scores between positive pairs in both retrieval settings (text and semantic). As depicted in Figure~\ref{fig:cosineimage}, our method exhibits a clear rightward shift in the distribution compared to baselines. This shift indicates a higher proportion of strongly aligned samples, confirming that SPANER achieves not just higher average similarity but also a more uniformly compact representation space.
These findings further validate the effectiveness of our shared prompt mechanism, which provides a common semantic scaffold for both modalities. By attending to the same set of learnable conceptual tokens, image and text features are drawn toward a mutual representational anchor, promoting tighter alignment. Unlike methods that rely solely on modality-specific encoders and instance-level contrastive learning, SPANER introduces structural guidance that supports deeper semantic grounding.
Importantly, our results challenge the view that contrastive learning inherently leads to modality separation or \textit{cone collapse} \cite{liang2022mind, fahim2024s}. Rather, we show that architectural design, specifically shared semantic anchors and coordinated attention, can mitigate this effect and enable deeper cross-modal integration.

\section{Extending to Audio}
We now demonstrate how SPANER can be extended to accommodate the audio modality without modifying the core architecture. Following the multimodal alignment setup of \cite{lin2023crossmodal}, we incorporate an audio encoder based on EsResNeXt \cite{guzhov2021esresne}, which has been adopted in audio-visual retrieval benchmarks. We evaluate this extension on two recently proposed datasets, ImageNet-ESC-19 and ImageNet-ESC-27 \cite{lin2023crossmodal}, which align environmental sounds with corresponding visual categories. Through a series of retrieval experiments, we show that SPANER can embed audio features into the same shared semantic space as vision and text, achieving competitive performance with prior baselines, and validating its modular and extensible design.
The following subsections outline the integration approach, experimental setup, and key findings.

\subsubsection{EsResNeXT into existing SPANER}
Following \cite{lin2023crossmodal}, we adopt the same audio backbone used in AudioCLIP \cite{guzhov2022audioclip}. 
While AudioCLIP pairs EsResNeXt with the RN50 variant of CLIP, we instead use EsResNeXt \cite{guzhov2021esresne} as a standalone audio encoder in our setup. 
As with section~\nameref{sec:vl}, our framework is agnostic to the specific choice of encoder; any audio feature extractor could be used, provided it outputs fixed-length representations. 
We select EsResNeXt to enable a fair and direct comparison with prior baselines that also rely on this encoder.
EsResNeXt follows a CNN-based architecture tailored for audio representation learning. 
To extract audio representations \(x^{\text{audio}}\), raw audio waveforms are first converted into log-mel spectrograms and segmented into overlapping patches \([a_1, a_2, \ldots, a_n]\), each representing a local time-frequency region. These patches are passed through a stack of convolutional layers, producing intermediate feature maps \([o_1, o_2, \ldots, o_n]\). 
To obtain a fixed-length representation for downstream alignment, we apply a temporal max pooling across the feature dimension:
    $x^{\text{audio}} = \text{MaxPool}([o_1, o_2, \ldots, o_n])$.
\(x^{\text{audio}}\) then serves as input to the audio branch of SPANER, where it is aligned with the shared semantic prompt and fused via the modality-specific cross-attention aligner.
\subsubsection{Experimental Setup}
To evaluate our audio extension, we follow the experimental protocol established by \cite{lin2023crossmodal}, using two datasets: \textbf{ImageNet-ESC-27}, which contains loosely matched image-sound pairs with partial label overlap, and \textbf{ImageNet-ESC-19}, a refined subset comprising more semantically consistent pairings. These benchmarks are derived by aligning label spaces between ImageNet-1K \cite{deng2009imagenet} and ESC-50 \cite{piczak2015dataset}, a well-known environmental sound classification dataset.
To extend our vision-language framework SPANER to include the audio modality, we adopt the configuration shown in Figure~\ref{fig:audio} (see Appendix for full illustration).
A new trainable CA Aligner is specifically initialised for the audio branch, while all other parts of the network are kept frozen.
Due to the dimensional mismatch between the audio features produced by the EsResNeXT encoder, \(x^{\text{audio}}\), and the already initialised shared prompt used in SPANER (Section~\nameref{sec:vl}), we apply a learnable linear projection to align their dimensions.
During training, each audio input is paired with a randomly sampled image from the 16-shot image pool used in the prior image-text alignment stage.
We use a batch size of 32 for ImageNet-ESC-27 and 16 for ImageNet-ESC-19, with all other training settings kept consistent with Section~\nameref{sec:vl}.
To assess effectiveness, we evaluate the audio extension using semantic retrieval accuracy and embedding distance metrics, which together capture both semantic matching and representational coherence. These results highlight SPANER’s ability to incorporate new modalities without altering the core architecture. Additional analysis are provided in the Appendix.

\subsubsection{Audio Retrieval Performance}

\begin{table}[]
    \centering
    \begin{tabular}{l|c|cc}
         \hline
         \textbf{Dataset} & \textbf{Method} & \textbf{1-shot} & \textbf{4-shot} \\
         \hline
         ImageNet-ESC-19 & Cross-Modal & 35.7 & 51.6 \\
         & \textbf{SPANER} (Ours) & 36.0 & 55.4 \\
         \hline
         ImageNet-ESC-27 & Cross-Modal & 35.0 & 48.5 \\
        & \textbf{SPANER} (Ours) & 33.0 & 50.5 \\
        \hline
    \end{tabular}
    \caption{Classification accuracy (\%) on audio datasets using k-shot training.
    Although our method is not exposed to the language modality during training, it achieves competitive performance through audio-semantic retrieval.
    }
    \label{tab:audioclass}
\end{table}

To assess the effectiveness of SPANER in extending to a new modality, we first evaluate audio-to-semantic retrieval on ImageNet-ESC-19 and ImageNet-ESC-27. 
As shown in Table~\ref{tab:audioclass}, our method achieves performance comparable to the previous Cross-Modal baseline \cite{lin2023crossmodal} in the 1-shot setting. 
With only a single support example, our SPANER achieves strong alignment without being exposed to language inputs during training, relying instead on the shared semantic space anchored by visual representations.
In the 4-shot setting, performance improves notably across both datasets, with SPANER outperforming the baseline. 
This suggests that with a modest number of audio examples, our framework can effectively adapt and incorporate audio into the shared semantic space. 
ESC-27, with its looser mappings, remains more challenging overall, as reflected in the lower retrieval scores for both baseline and our method. Nevertheless, SPANER maintains robustness even under these noisier conditions, suggesting that shared prompt grounding provides a stable foundation for cross-modal generalisation.


\subsubsection{Cosine Similarity and Semantic Tightness}

\begin{figure}[tbp]
    \centerline{\includegraphics[width=\linewidth]{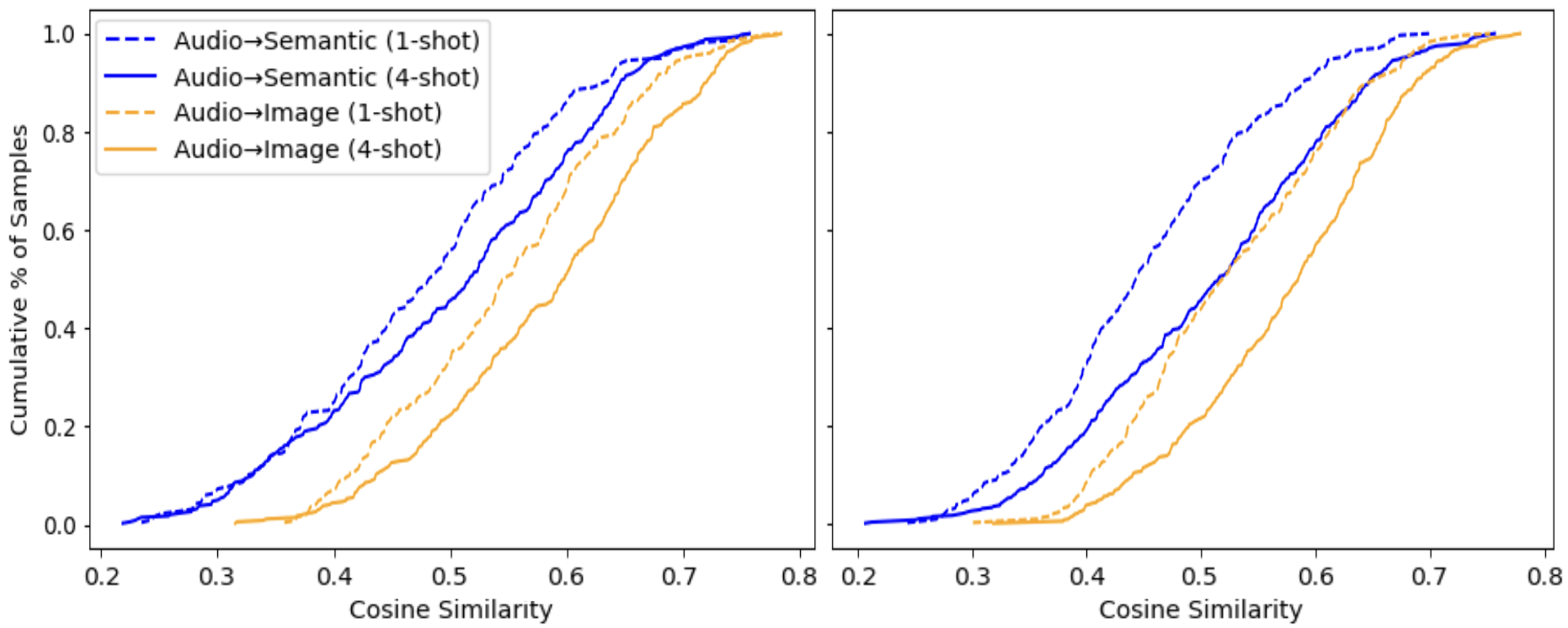}}
    \caption{CDF of pairwise cosine similarity scores for positive samples. Left: ImageNet-ESC-19. Right: ImageNet-ESC-27. Each plot compares audio-to-semantic and audio-to-image alignment under 1-shot and 4-shot training. Our method shows improved alignment with increasing shots, with audio-to-image generally yielding higher scores than audio-to-semantic due to its direct training objective.
}
    \label{fig:cosineaudio}
    \end{figure}


To quantitatively assess alignment tightness, we analyse cosine similarity distributions between audio and other modalities using CDFs. As shown in Figure~\ref{fig:cosineaudio}, similarity scores improve with increased training shots. Moreover, audio-to-image alignment consistently yields higher cosine similarity than audio-to-semantic, reflecting the direct supervision provided by the image branch during training.
While our audio-to-image similarity scores reach averages around 0.55 in the 4-shot setting, audio-to-semantic scores remain slightly lower, around 0.48. These values, while meaningful, are lower than those observed in the vision-language experiments (Table~\ref{tab:cosineimage}), where pre-aligned encoders like CLIP facilitated much tighter clustering.
We attribute this gap to two primary factors: (1) the limited number of examples available for learning audio embeddings, and (2) the lack of pretrained alignment between EsResNeXt and CLIP-like text encoders, which places greater pressure on SPANER’s shared prompt to bridge modalities indirectly. Despite this, the model achieves encouraging alignment quality, demonstrating the framework’s ability to generalise through structure and supervision rather than architectural modifications.




\section{Ablation Studies}
    To better understand the design choices in SPANER, we conduct a series of ablation experiments. 
These studies examine the effects of shared prompt length, fusion mechanisms, text template strategies, and projection dimensions on both image and audio retrieval tasks. 
By isolating these components, we provide insight into how different factors influence semantic alignment quality and generalisation across modalities. 
The following subsections present our findings.

\begin{figure}[tbp]
    \centering
    \includegraphics[width=\linewidth]{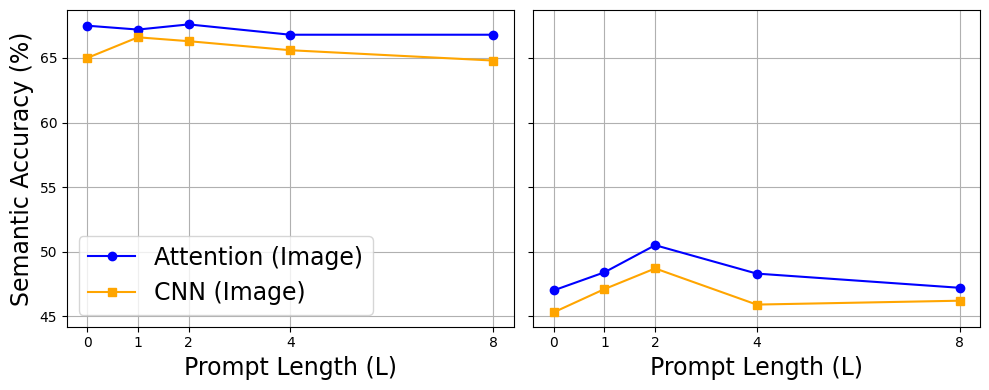}
    \caption{Impact of prompt length and fusion mechanism on semantic retrieval performance. Results for left plot correspond to image-semantic retrieval on ImageNet, and right plot refers to audio-semantic retrieval on ImageNet-ESC-27. While prompt length shows minimal effect in the two-modality (vision-language) setup, its role becomes more critical with the addition of a third modality (audio), emphasising the need for a balanced shared-prompt design.
    }
    \label{fig:promptfusion}
\end{figure}
\begin{figure}
    \centering
    \includegraphics[width=\linewidth]{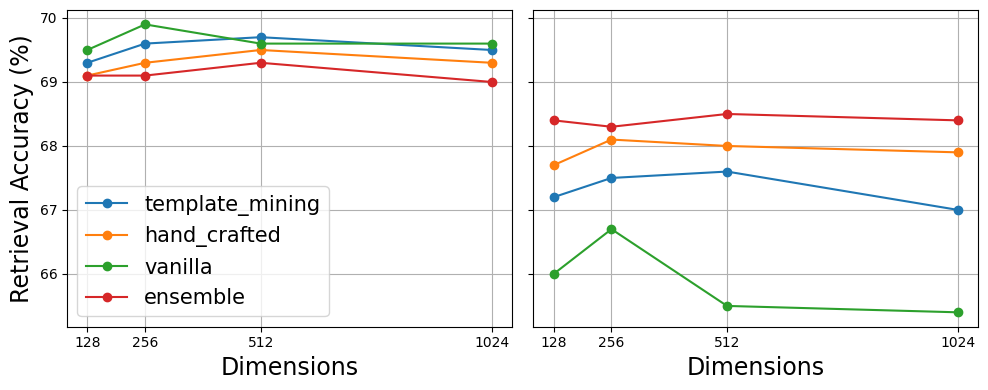}
    \caption{Retrieval accuracy on ImageNet across different text template strategies and shared space projection dimensions. Left plot: image-to-text retrieval accuracy (\%). Right plot: semantic retrieval accuracy (\%)}
    \label{fig:textaug_proj}

\end{figure}
\subsubsection{Shared Prompt in Modality Extension} 
To evaluate the impact of fusion strategies and shared prompt length, we compare attention-based \cite{vaswani2017attention} and CNN(Hyena)-based \cite{poli2023hyena} aligners across varying prompt lengths \(L\) (Figure~\ref{fig:promptfusion}), measuring semantic retrieval performance for both image and audio modalities. 
Specifically, in the attention-based variant, we use the mechanism defined in Equation~\ref{eq:attn_outs}, and we switch this with CNN over the concatenated shared and input tokens. 
Under the VL setting, image-semantic retrieval (left plot) on ImageNet remains relatively stable regardless of prompt length or fusion type, suggesting limited sensitivity when aligning only two modalities.
However, when extending SPANER to audio and evaluating audio-semantic retrieval (right plot) on ImageNet-ESC-27, the role of the shared prompt becomes more pronounced. 
Moderate prompt lengths (e.g., \(L=2\)) yield the best results--particularly for the attention-based aligner--indicating that a compact shared prompt is sufficient to encode useful cross-modal context. 
Notably, excessively long prompts tend to degrade performance. 
We attribute this to the fact that the final representation \(f\) is derived from a single summary token, and an overly long prompt may introduce noisy or redundant information that dilutes the semantic signal. 
Data augmentation can influence downstream retrieval performance \cite{lin2023crossmodal}. 
In this experiment, we investigate the impact of different text template strategies and projection dimensions on both image-to-text and semantic retrieval, as shown in Figure~\ref{fig:textaug_proj}.
The vanilla template uses a fixed format (``a photo of a [CLASSNAME]''), which provides consistent target phrases and achieves the highest text retrieval accuracy (left plot). 
However, this simplicity limits its semantic generalisation, resulting in comparatively lower performance in semantic retrieval (right plot). 
On the other hand, the ensemble strategy employs 180 diverse templates, capturing a broad range of textual variations. 
This leads to lower image-to-text retrieval scores due to reduced alignment specificity, but yields the best semantic retrieval performance--highlighting the trade-off between retrieval precision and semantic robustness.
To balance this, we adopt the template mining approach (21 templates), which offers a strong compromise between the two extremes. 
For consistency, all previous experiments use this strategy.
All template variants, are sourced from the dataset construction methodology proposed in \cite{lin2023crossmodal}.
We also evaluate the effect of varying the final projection dimension used in the shared embedding space. Notably, for the template mining setup, performance peaks at 512 dimensions--matching the original CLIP backbone projection size--therefore we adopt this as the default throughout our framework.
Lastly, we also experimented with common image augmentations but observed no significant performance gains. 
As a result, we provide only two views per image during training: one random crop and one center crop.
    
\section{Conclusion}
    In this work, we introduced Shared Prompt AligNER (SPANER), a modality-agnostic framework for aligning diverse inputs into a unified semantic space using shared prompts. 
Rather than optimising solely for classification or retrieval accuracy, SPANER focuses on semantic coherence across modalities--enabling generalisation to new data types such as audio without requiring retraining or architectural modifications. 
Our experiments across vision-language and audio benchmarks demonstrate that SPANER effectively bridges modality gaps while maintaining performance in cross-modal and semantic retrieval tasks. Overall, our findings highlight the potential of shared prompts as a scalable and modular approach for multimodal understanding, while also pointing toward future directions involving adaptive prompts, improved fusion strategies, and domain-aware alignment techniques.




\bibliography{references}

\newpage

\appendix
\clearpage
    \section{Algorithm} 
\label{appendix:algorithm}
\begin{algorithm}[]
    \caption{Contrastive Loss for Paired Representations}
    \label{alg:contrastive_loss}
    \begin{algorithmic}[1]
    \REQUIRE Paired feature matrices $z^1, z^2 \in \mathbb{R}^{B \times d}$
    \STATE $z^1 \gets \text{Normalise}(z^1)$
    \STATE $z^2 \gets \text{Normalise}(z^2)$
    \STATE $M \gets z^1 \cdot z^{2\top}$ \hfill \textit{// Cosine similarity matrix}
    \STATE $\text{targets} \gets [0, 1, \dots, B{-}1]$
    \STATE $\mathcal{L} \gets \text{CrossEntropy}(M, \text{targets})$
    \RETURN $\mathcal{L}$
    \end{algorithmic}
\end{algorithm}

\begin{algorithm}[]
\caption{SPANER Training Loop for newly initialised}
\label{alg:shared_prompt_training}
\begin{algorithmic}[1]
\REQUIRE Batch of paired modality embeddings $\{x^1, x^2\}$, shared prompt tokens $S = \{s_1, \dots, s_n\}$, temperature-scaled contrastive loss $\mathcal{L}_{\text{LOSS}}$, balancing coefficient $\lambda$
\FOR{each batch in dataset}
    \STATE $[x^{1\prime}; S'] \gets \text{Attn}^{1}([x^1; S])$
    \STATE $[x^{2\prime}; S'] \gets \text{Attn}^{2}([x^2; S])$
    \STATE $z^1 \gets \text{MaxPool}(x^{1\prime}, S')$
    \STATE $z^2 \gets \text{MaxPool}(x^{2\prime}, S')$
    \STATE $\mathcal{L}_{\text{ca}} \gets \text{LOSS}(z^1, z^2)$ \hfill \textit{// Algorithm 1}
    \STATE $f^1 \gets \text{Proj}^1(x^1)$
    \STATE $f^2 \gets \text{Proj}^2(x^2)$
    \STATE $\mathcal{L}_{\text{align}} \gets \text{LOSS}(f^1, f^2)$
    \hfill \textit{// Algorithm \ref{alg:contrastive_loss}}
    \STATE $\mathcal{L} \gets \mathcal{L}_{\text{align}} + \lambda \cdot \mathcal{L}_{\text{ca}}$
    \STATE Update parameters using gradients from $\mathcal{L}$
\ENDFOR
\end{algorithmic}
\end{algorithm}

\section{Vision Language - Further Analysis}
\subsection{Generalisation and the Semantic-Compactness}
While our method demonstrates strong alignment within the ImageNet domain, it is essential to evaluate its robustness to distributional shifts, an growing concern in multimodal learning. To this end, we conduct experiments on multiple ImageNet variants: ImageNet-V2, -Sketch, -A, and -R. These datasets introduce various types of shifts, including changes in texture, style, abstraction, and content composition. They serve as standard benchmarks for evaluating generalisation in vision-language models.
\begin{table}[ht]
    \centering
    \resizebox{\linewidth}{!}{
    \begin{tabular}{l|ccccc}
    \hline
    \textbf{Method} & \textbf{ImageNet} & \textbf{-V2} & \textbf{-Sketch} & \textbf{-A} & \textbf{-R} \\
    \hline
    ZS CLIP & 66.7 & 60.8 & 46.2 & 47.8 & 74.0 \\
    CoOp           & 71.5 & 64.2 & 47.8 & 49.7 & 75.2 \\
    CLIP-Adapter   & 71.1 & 61.8 & 42.4 & 42.3 & 67.7 \\
    CoCoOp         & 71.0 & 64.1 & 48.8 & 50.6 & 76.2 \\
    MaPLe          & 70.7 & 64.1 & 49.2 & 50.9 & 76.9 \\
    MMA            & 71.0 & 64.3 & 49.1 & 51.1 & 77.3 \\
    \textbf{SPANER} (Ours)           & 69.7 & 60.9 & 42.3 & 41.8 & 67.0 \\
    \hline
    \end{tabular}
    }
    \caption{Domain generalisation results (text retrieval accuracy \%) on ImageNet and its distributional shifts.}
    \label{tab:shift}
\end{table}

As seen in Table \ref{tab:shift}, SPANER performs competitively on the original ImageNet test set (69.7\%), but exhibits noticeable performance drops across the shifted datasets. On ImageNet-V2 and -R, our model underperforms most baselines. The performance drop is even more pronounced on ImageNet-Sketch and -A, which contain stylised and adversarial samples, respectively. These results highlight a potential trade-off introduced by our architecture. While SPANER yields highly aligned and semantically coherent embeddings, this compactness may come at the cost of adaptability to unseen or stylistically varied domains.

It is worth noting, however, that domain generalisation is not the primary objective of SPANER. Our method is optimised for achieving compact and coherent multimodal alignment in a shared semantic space. The observed limitations under distributional shift suggest that high alignment quality and domain robustness may be at odds in certain architectural settings. Striking the right balance between semantic compactness and representational elasticity remains a key challenge for future work. One possible future direction is to incorporate adaptive prompt modulation strategies or instance-aware alignment objectives that retain the benefits of a shared conceptual scaffold while selectively relaxing alignment constraints in response to distributional variance.

\subsection{Confusion Patterns Between Semantically Similar Classes}
\begin{figure*}[tbp]
    \centering
    \includegraphics[width=\linewidth]{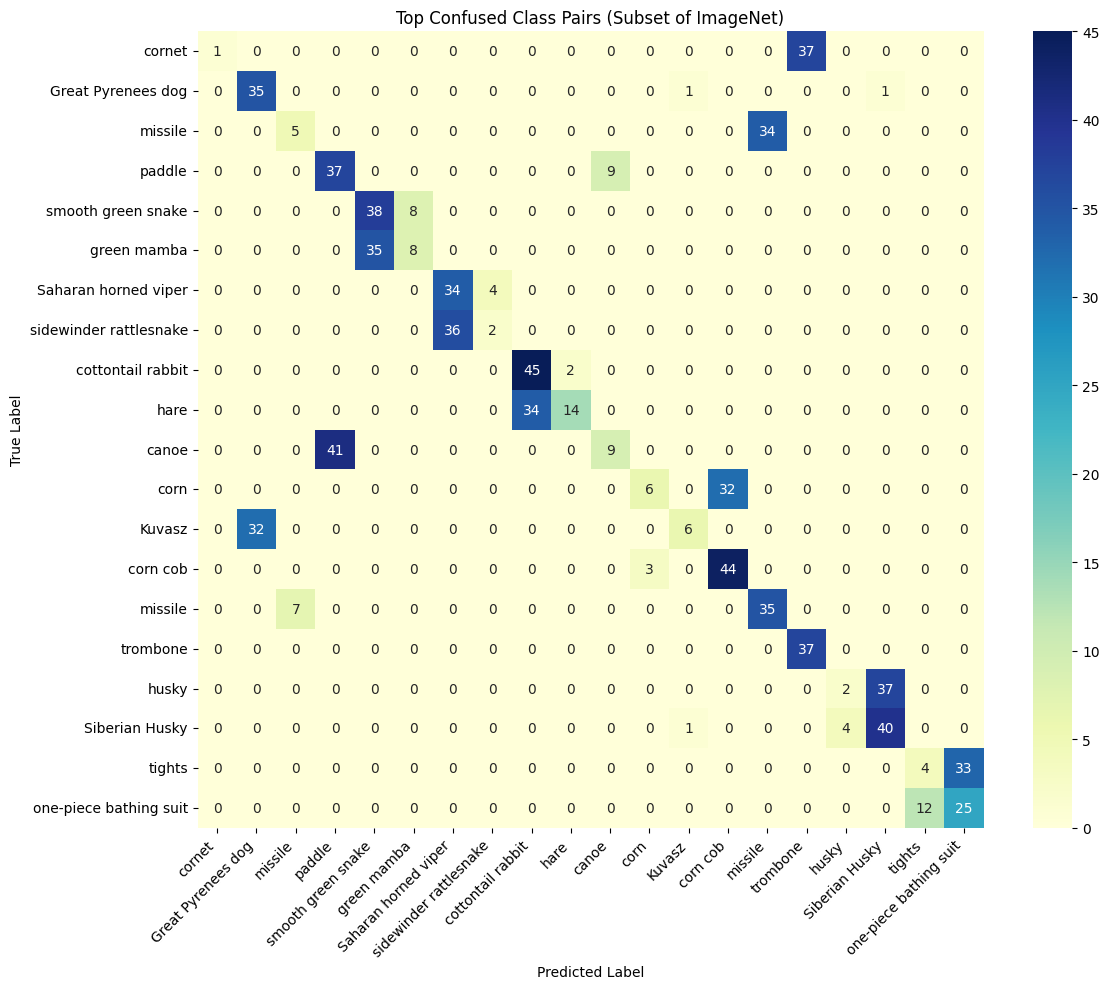}
    \caption{Subset of the confusion matrix showing the 20 most frequent incorrect semantic retrievals on the ImageNet dataset. Although SPANER maps instances into a shared semantic space, retrieval errors still occur among semantically or visually similar categories--such as dog breeds, snake species, or musical instruments—highlighting the challenge of distinguishing fine--grained concepts even within a semantically aligned embedding.}
    \label{fig:confusion_heatmap}
\end{figure*}

We conduct a qualitative analysis to explore the nature of retrieval errors made by our model. Specifically, we examine failure cases in semantic retrieval and identify common patterns of confusion between closely related categories. This analysis offers further insight into how SPANER operates when distinguishing fine-grained semantic concepts within densely populated regions of the embedding space.
We observe that the majority of incorrect retrievals are not due to arbitrary mismatches but rather occur between semantically or visually similar classes. For example, the model may confuse different dog breeds such as “Husky” and “Siberian Husky,” or closely related types like “tights” and “one-piece bathing suit” These errors suggest that while the model effectively captures high-level semantic groupings, it may struggle with fine-grained distinctions within those groupings, particularly when the visual features are highly similar or the textual labels share overlapping conceptual attributes.
This pattern aligns with our hypothesis that SPANER’s shared prompt mechanism encourages representations to cluster around semantically meaningful anchors. While this promotes strong alignment for broad semantic categories, it may also lead to tighter grouping of instances that should ideally remain distinguishable at a finer granularity. This behaviour reflects a double-edged property of semantic compactness: it improves conceptual cohesion but can obscure class-specific nuances.
To better understand these confusion dynamics, we compute a class-wise semantic retrieval confusion matrix over the ImageNet validation set. Each entry in the matrix represents the frequency with which a query from class $c_i$ retrieves an instance from class $c_j$. We then extract the most frequently confused class pairs and visualise them in a subset heatmap, focusing on the top 20 classes with the highest misalignment rates. A representative portion of this confusion map is provided in Figure~\ref{fig:confusion_heatmap}.

\section{Audio - Further analysis}

\subsection{Cross-Modal Retrieval: from Semantics}
\begin{table}[]
    \centering
    \begin{tabular}{l|c|cc}
        \hline
         \textbf{Dataset} & \textbf{Retrieval} & \textbf{1-shot} & \textbf{4-shot} \\
         \hline
         ImageNet-ESC-19 & Semantic\(\rightarrow\)Audio & 57.9 & 84.2 \\ 
         & Image\(\rightarrow\)Audio & 62.2 & 86.8 \\
         \hline
         ImageNet-ESC-27 & Semantic\(\rightarrow\)Audio & 63.0 & 85.2 \\
         & Image\(\rightarrow\)Audio & 60.5 & 78.9 \\
         \hline
    \end{tabular}
    \caption{Retrieval accuracy (\%) on audio benchmarks using 1-shot and 4-shot training.
    Retrieval is performed from either semantic (class name) or image queries to audio.
    Despite audio being trained only with visual inputs, our model enables strong cross-modal retrieval performance, demonstrating the effectiveness of shared semantic grounding.}
    \label{tab:toaudio}
\end{table}


We further assess SPANER’s ability to retrieve audio samples given either a semantic label (class name) or image input as the query. Results are presented in Table~\ref{tab:toaudio}. Across both benchmarks and both retrieval directions, performance increases substantially with additional support examples. This again highlights SPANER’s ability to learn effective alignments even with limited data.

Interestingly, we observe that retrieval from image or semantic queries to audio tends to outperform the reverse direction (audio-to-image or audio-to-semantic). This asymmetry is expected, given that high-resource modalities, vision and text, primarily shape the shared semantic space. At the same time, the audio pathway is newly introduced and aligned only via cross-modal supervision with vision. The audio features, originating from a standalone CNN-based encoder, exhibit higher variability and weaker grounding than the text- or vision-derived embeddings. Nonetheless, the high accuracy achieved in both retrieval tasks confirms that the shared prompt mechanism provides sufficient guidance for meaningful alignment.

\subsection{Audio Retrieval under Low Resource Settings}
\begin{table*}[htbp]
\centering
\begin{tabular}{p{4cm}|p{11cm}}
\toprule
\textbf{Ground Truth} & \textbf{Top-5 Predictions} \\
\midrule
\multicolumn{2}{c}{\textbf{Correct Top-1 Predictions}} \\
\midrule
\texttt{church}         & \texttt{church}, \texttt{computer mouse}, \texttt{can opener}, \texttt{high-speed train}, \texttt{Otterhound} \\
\texttt{rooster}        & \texttt{rooster}, \texttt{hen}, \texttt{airliner}, \texttt{tree frog}, \texttt{water bottle} \\
\texttt{sandbar}        & \texttt{sandbar}, \texttt{Otterhound}, \texttt{fire screen}, \texttt{Egyptian Mau}, \texttt{water bottle} \\
\texttt{airliner}       & \texttt{airliner}, \texttt{Otterhound}, \texttt{fly}, \texttt{computer keyboard}, \texttt{chickadee} \\
\texttt{cricket insect} & \texttt{cricket insect}, \texttt{chickadee}, \texttt{tree frog}, \texttt{water jug}, \texttt{church} \\
\midrule
\multicolumn{2}{c}{\textbf{Incorrect Top-1 Predictions}} \\
\midrule
\texttt{water jug}      & \texttt{sink}, \texttt{Egyptian Mau}, \texttt{computer keyboard}, \texttt{bighorn sheep}, \texttt{hen} \\
\texttt{Otterhound}     & \texttt{digital clock}, \texttt{computer keyboard}, \texttt{fly}, \texttt{computer mouse}, \texttt{Egyptian Mau} \\
\texttt{fly}            & \texttt{fire screen}, \texttt{chickadee}, \texttt{airliner}, \texttt{church}, \texttt{digital clock} \\
\texttt{pig}            & \texttt{water bottle}, \texttt{water jug}, \texttt{rooster}, \texttt{cricket insect}, \texttt{hen} \\
\texttt{sink}           & \texttt{Otterhound}, \texttt{can opener}, \texttt{computer mouse}, \texttt{high-speed train}, \texttt{fire screen} \\
\bottomrule
\end{tabular}
\caption{Qualitative analysis of audio semantic retrieval: 5 correct and 5 incorrect Top-1 predictions with their Top-5 semantic retrievals.}
\label{tab:qual_audio_preds}
\end{table*}

Beyond quantitative evaluation, we perform a qualitative analysis of audio-to-semantic retrieval on the ImageNet-ESC-27 benchmark. Table~\ref{tab:qual_audio_preds} presents five examples where the correct label was retrieved at top-1, and five where the top-1 prediction was incorrect.

While the model occasionally retrieves acoustically or semantically related categories (e.g., \texttt{rooster} alongside \texttt{hen}, or \texttt{water jug} with \texttt{sink}), many top-5 predictions include irrelevant or incoherent labels (e.g., \texttt{digital clock}, \texttt{computer mouse}, or \texttt{Otterhound} for non-speech sounds), indicating inconsistencies in the shared embedding space across modalities.

We attribute these issues to two primary factors: (1) the backbone lacks strong prealignment between audio and semantic modalities--unlike CLIP, which already benefitted from large-scale vision-language pretraining, audio is only introduced later; and (2) the ImageNet-ESC-27 dataset is relatively small and imbalanced, limiting the diversity and generalisability of audio representations. These limitations result in noisy or semantically irrelevant retrievals, highlighting the need for improved multimodal alignment methods--particularly in low-resource or under-aligned modality settings. This represents a promising direction for future research.

\begin{figure}
        \centering
        \includegraphics[width=0.7\linewidth]
        {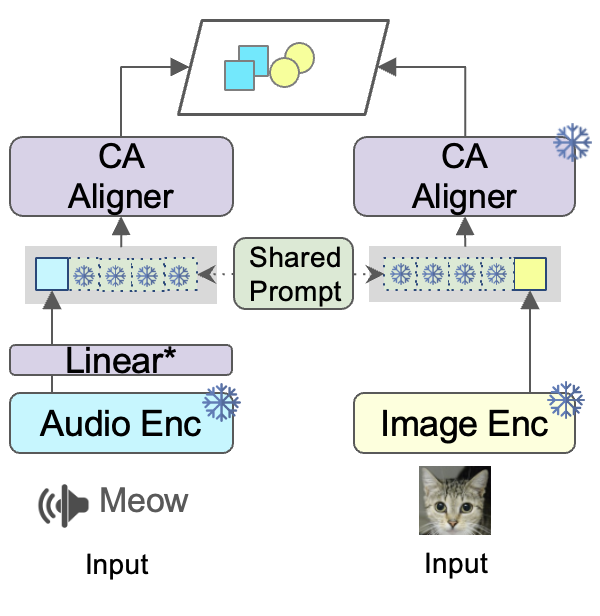}
        \caption{Extending SPANER to the audio modality. The trained shared prompt and its corresponding CA aligner are kept frozen. A newly initialised CA aligner is introduced for the audio branch, which is trained while the rest of the system remains unchanged. A learnable linear projection is applied to the audio encoder output if there is a dimensional mismatch with the shared prompt.}
        \label{fig:audio}
\end{figure}


\end{document}